\def\year{2019}\relax
\documentclass[letterpaper]{article} 
\usepackage{aaai19}  
\usepackage{times}  
\usepackage{helvet}  
\usepackage{courier}  
\usepackage{url}  
\usepackage{graphicx}  
\usepackage{amsmath}
\usepackage{amssymb}
\usepackage{multirow}
\begin{document}
%
\title{Memory-Augmented Temporal Dynamic Learning for Action Recognition}
\author{Yuan Yuan, Dong Wang, Qi Wang\\
School of Computer Science and Center for OPTical IMagery Analysis and Learning (OPTIMAL), \\
Northwestern Polytechnical University, Xi'an 710072, China,\\
\{y.yuan1.ieee, nwpuwangdong, crabwq\}@gmail.com}
\maketitle
\begin{abstract}
Human actions captured in video sequences contain two crucial factors for action recognition, i.e., visual appearance and motion dynamics. To model these two aspects, Convolutional and Recurrent Neural Networks (CNNs and RNNs) are adopted in most existing successful methods for recognizing actions. However, CNN based methods are limited in modeling long-term motion dynamics. RNNs are able to learn temporal motion dynamics but lack effective ways to tackle unsteady dynamics in long-duration motion. In this work, we propose a memory-augmented temporal dynamic learning network, which learns to write the most evident information into an external memory module and ignore irrelevant ones. In particular, we present a differential memory controller to make a discrete decision on whether the external memory module should be updated with current feature. The discrete memory controller takes in the memory history, context embedding and current feature as inputs and controls information flow into the external memory module. Additionally, we train this discrete memory controller using straight-through estimator. We evaluate this end-to-end system on benchmark datasets (UCF101 and HMDB51) of human action recognition. The experimental results show consistent improvements on both datasets over prior works and our baselines.
\end{abstract}

\frenchspacing

\section{Introduction}
In recent years, video based human action recognition has received increasing attention from the research community \cite{wang2011action,simonyan2014two,tran2015learning,carreira2017quo}, owing to its great potential value in many real-world applications like surveillance \cite{lin2008human}, abnormal activity detection \cite{boiman2007detecting} and so forth. Unlike recognition in static images, a distinctive aspect of the action recognition in videos is the motion dynamics, which is a crucial factor in addition to visual appearance. The performance of action recognition system depends, to a large extent, on whether the dynamics can be effectively represented and utilized. A key goal of this work is to enhance the model's capacity for learning and capturing the motion dynamics in videos.

Encouraged by the success of Convolutional Neural Network (CNNs) on computer vision tasks such as image classification, many researchers have adopted similar methods for video understanding and human action recognition \cite{simonyan2014two,feichtenhofer2016convolutional,zha2015exploiting,sun2015human} and achieved remarkable performance on public benchmark datasets. Deep CNNs have been shown to have an extraordinary capacity for learning discriminative representation from raw visual data. The most successful action recognition algorithm, two-stream convents, utilizes two independent CNNs to extract pre-frame features from RGB and optical flow images, followed by simple or strategic pooling across the temporal domain. However, traditional two-stream convnets learn the action representation frame by frame and lack the capacity to model long-range dynamics across the temporal domain. Recurrent Neural Networks (RNNs) is a straightforward choice to exploit the sequential structure in videos. Long Short-Term Memory (LSTM) and its variants have been explored to incorporate frame-level features in several works \cite{xingjian2015convolutional,li2018videolstm,ma2017ts}. These models tend to work well for actions with short duration and little movement. However, these approaches would incur low accuracy when applied to long-range and complex motions, as verified by our experiments. These challenges motivate us to design an effective and efficient module to (1) capture salient motion dynamics in long-range temporal structure and (2) compose complex motion information across time for distinguishing actions.

In this paper, we propose a novel network structure, named memory-augmented temporal dynamic learning network, where the frame-level feature representation is stored and recalled from an external memory module, to maintain the informative motion dynamics for classifying actions. It can more effectively learn the long-term motion dynamic without increasing the model complexity. Specifically, a video is converted to a sequence of frame-level features by CNN and communicates with external memory according to a memory controller. There are three elements fed into memory controller at every frame, i.e., memory history, context embedding and current frame-level feature. The memory history is constructed by all features in the memory module, and the context embedding is obtained using a Long Short-Term Memory network that skilled in modeling short-range motion dynamics. We introduce a threshold-based selection mechanism to make memory controller output a discrete decision on whether write the current frame-level feature into memory module. After processed every frame features, the final action representation for recognizing actions is obtained by applying average pooling to features in the memory module.  At the same time, the straight-through estimator is utilized to make the discrete memory controller trainable in end-to-end manner. We evaluate our model on two benchmark datasets, and outperform our baseline and state-of-the-art performance.

The main contribution of this paper is summarized as follows:
\begin{itemize}
\item We design a memory-augmented temporal dynamic learning network for action recognition. An external memory module is attached to CNNs to store salient and informative motion dynamics in videos, and greatly enhances capacity to encode long-term and complex motion dynamics in long-range temporal structure.
\item We propose a discrete memory controller, that takes in the memory history, context embedding and current feature as inputs, to control the writing process of the external memory module. This allows efficiently leveraging information from different temporal scale and improves the representation power, as demonstrated by our experiments.
\item We extensively evaluate our algorithm on large-scale datasets UCF101 and HMDB51. Our method performs favorably against state-of-the-art action recognition methods and our baseline.
\end{itemize}

\section{Related Work}
Motivated by the impressive performance of CNNs on image classification \cite{krizhevsky2012imagenet}, semantic segmentation \cite{long2015fully} and other computer vision work \cite{8418840}, several recent works have utilized CNN-based architectures for video based human action recognition. Karpathy et al. \cite{karpathy2014large} directly apply CNNs to extract frame-level features and exploit multiple simple temporal pooling methods, including early fusion, late fusion, and slow fusion. But these approaches only yield a modest improvement over single frame baseline, indicating that motion dynamic information is hard to model by directly pooling spatial features from CNNs.

In view of this, Tran et al. \cite{tran2015learning} propose Convolutional 3D (C3D) and construct a deep C3D neural network with 3D convolution filters and 3D pooling layers, which operate on short video clips over space and time simultaneously. Noticing that 3D kernels only cover a short range of the sequence when filtering video clips, Simonyan et al. \cite{simonyan2014two} incorporate motion information by training a temporal stream of CNN on optical frames in addition to spatial stream with RGB frame input. By simple fusing probability scores from these two-stream CNNs, the accuracy of action recognition is significantly boosted. Moreover, several attempts have been made to learn subtle spatio-temporal relationships between appearance and motion in order to improve recognition accuracy. Feichtenhofer et al. \cite{feichtenhofer2016convolutional} study a number of ways of integrating two-stream CNNs spatially and temporally. They propose s spatiotemporal fusion method by generalizing residual networks. Wang et al. \cite{feichtenhofer2017multiplier} introduce the spatiotemporal compact bilinear operator to efficiently fuse spatial and temporal features hierarchically. However, the only several consecutive optical flow frames are fed into temporal stream CNN, so that it cannot capture longer-term motion patterns associated with certain human actions.

To enable the model to learn long-term motion dynamic, Ng et al. \cite{yue2015beyond} take advantage of LSTM to fuse features across a longer temporal range. However, the vanilla LSTM model is not satisfactory for learning the spatial correlations and motion dynamics between the frames. \cite{xingjian2015convolutional} extends LSTM to ConvLSTM, which replace linear multiplicative operation with spatial 2D convolution, so that it can learn the spatial patterns along the temporal domain. Furthermore, Shikhar et al. \cite{sharma2015action} implant soft attention module in LSTM to learn which parts in the frames are fatal for the task at hand and assign higher importance to them. Spring from the soft-Attention LSTM, VideoLSTM \cite{li2018videolstm} hardwire convolutions and introduce motion-based attention to guides better the attention towards the relevant spatio-temporal locations. However, this complex architecture does not bring significant performance improvement. Recently, Sun et al. \cite{sun2017lattice} propose Lattice-LSTM (L$^2$TSM), which extends traditional LSTM by learning independent hidden state transition of memory cells for individual spatial locations. But it does not address the complex backgrounds similar scenes problem in different categories very well. Additionally, Ma et al. \cite{ma2017ts} study two different ways, i.e., Temporal Segment LSTM and Temporal Inception, to extract spatio-temporal information by systematically explored possible network architectures. Wang et al. \cite{wang2016temporal} propose a segmental network, which splits a long sequence into several segments followed by sparse sampling, and achieve state-of-the-art performances.

From a technical standpoint, our approach is based on the dynamic memory networks. Memory networks are typically used to tackle simple logical reasoning problem in natural language processing like question answering and sentiment analysis. The pioneering works Neural Turing Machine (NTM) \cite{graves2014neural} and Memory Neural Networks (MemNN) \cite{sukhbaatar2015end} both propose an addressable external memory with the reading and writing mechanism. For computer vision tasks, memory networks have been adopted in visual tracking \cite{yang2018learning}, video question answering \cite{gao2018motion} and so on. In recent work \cite{vu2018memory}, an external memory module is employed to learn long-term online video representation in recurrent manner. Moreover, the proposed method is similar to attention-based action recognition model, which enhances the evident information by assigning soft weights on frames of one video. Except for temporal soft attention mechanism in soft-Attention LSTM \cite{sharma2015action}, Long et al. \cite{long2018attention} propose an attention-based shifting operation to integrate informative local features and an multimodal keyless attention model is proposed to fuse visual and acoustic features for action recognition in \cite{long2018multimodal}

\begin{figure*}
\begin{center}
   \includegraphics[width=1.0\linewidth]{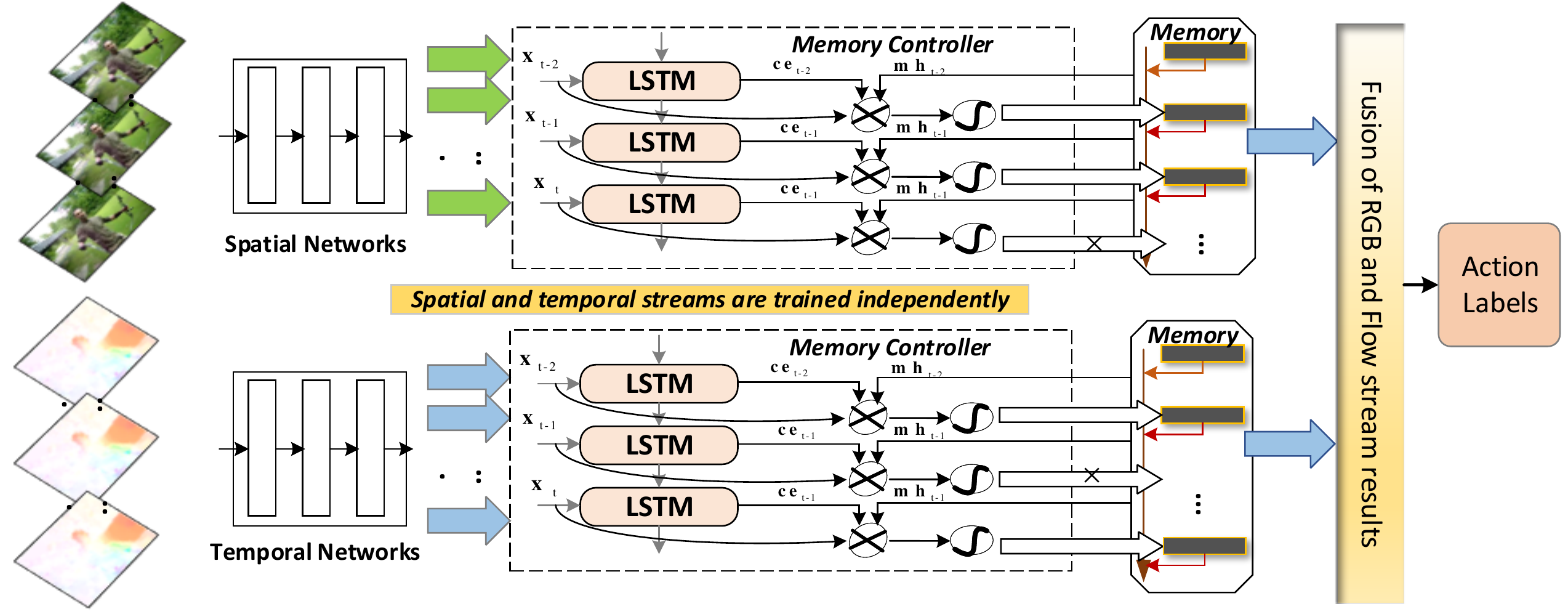}
\end{center}
     \caption{The overview of Memory-Augmented Temporal Dynamic Learning Network for action recognition. First, a set of video sequences are passing through CNN to extract convolutional features. Then, an LSTM unit is used to obtain context embedding by processing these features sequentially. Meanwhile, the memory items are recalled to constructed memory history. At each timestep, the memory controller takes these as inputs and outputs the discrete memory write decisions. Finally, the memory items in the memory module are pooled to recognizing action in the video. The final classification results are the average of spatial and temporal stream results.}
\label{Fig-overview}
\end{figure*}

\section{Approach}
The primary goal of this paper is to enhance the model's capacity for learning long-term and complex motion for action recognition in videos, by capturing salient and informative motion dynamics across the long-range temporal structure. A structured and growing external memory has been demonstrated is capable of learning long-term sequential pattern in \cite{joulin2015inferring}. Inspired by this idea, we propose to continuously read and write external memory module over time to obtain stable and discriminative representations of the human action captured by the video, which is verified by working with different CNN models.

\subsection{Overview}
We design a end-to-end dynamic memory network that writes task-relevant features into a growing external memory module according to a discrete memory controller. Figure \ref{Fig-overview} shows the overall architecture of our design. Specifically, given a whole video in the form of sparse sampled frame sequence with length $T$, suppose at each time $t$ with $t = 1, \ldots ,T$, the network first produces convolution features ${\bf{x}}_t$ with $d$ dimension for individual frames via arbitrary convolutional network. As shown in many CNN-based method \cite{wang2016temporal,simonyan2014two}, after training, these convolutional features can capture significant appearance evidences for action recognition. Subsequently, convolutional features ${\bf{x}}_{1,\ldots,t-1}$ are fed into a LSTM to obtain context embedding ${\bf{ce}}_t$ for each time $t$. Meanwhile, the memory history ${\bf{mh}}_t$ is calculated by current external memory module ${\bf{M}}_t$. At the end, the memory controller takes context embedding ${\bf{ce}}_t$, memory history ${\bf{mh}}_t$ and convolutional feature ${\bf{x}}_t$ as inputs and outputs a discrete decision ${\bf{s}}_t \in \{0,1\}$ on whether write convolutional feature ${\bf{x}}_t$ into external memory module ${\bf{M}}_t$. The combination of context embedding and memory history enables the model to capture informative motion dynamics in long-range temporal structure.

The whole input sequence is processed in a sequential manner. The memory history ${\bf{mh}}_{t+1}$ is obtained after external memory module ${\bf{M}}_t$ have been updated at time $t$, and the features in external memory module ${\bf{M}}_T$ are averaged to get final representation for action recognition. In what follows, we will explain the read and write process of external memory over time with details, as well as how the discrete decisions are made by the memory controller.

\subsection{Updating Memory Module over Time}
Different from previous works that either mixes memory with computation in the recurrent network or mimics the one-dimensional memory with elaborate access mechanism in the Turing machine/von Neumann architecture, we propose a growing memory module to process arbitrarily long sequences theoretically. This is intuitive because frame sequences that capture same human action may have different length with different sample rate. But more importantly, a growing memory module is suitable for model long-term and complex motion in videos. To be specific, our memory module is represented by a list $M$, with the length of it equals to the number of memory items and memory item is $D$ dimension feature vector $\bf{m}$. Memory module can be updated after processing each frame, thus we use ${\bf{M}}_t^{'}$ to represent memory before updating for the current frame so far and ${\bf{M}}_t$ denotes memory module after processing current frame, that is, ${\bf{M}}_t^{'}$ and ${\bf{M}}_{t-1}$ are the same.

\textbf{Memory Read and Write.} When processing the first frame, the external memory module is empty so that memory history ${\bf{mh}}_1$ is a zero feature vector. Suppose memory module ${\bf{M}}_t, t>1$ is not empty with length $N_t > 0$, and then ${\bf{mh}}_t$ is average of all memory items in memory module,
\begin{equation}
\label{eq-mh}
{\bf{mh}}_t = \sum\limits_{i = 1}^{{N_t}} {{{\bf{m}}_i}},
\end{equation}
which is a simple but effective memory read mechanism. Meanwhile, context embedding ${\bf{ce}}_t$ is calculated by a custom LSTM, which will be elaborated in next section. Next, the context embedding ${\bf{ce}}_t$, memory history ${\bf{mh}}_t$ and convolutional feature ${\bf{x}}_t$ are fed into the memory controller to obtain discrete decision ${\bf{s}}_t \in \{0,1\}$, and the memory module is updated as follows,
\begin{equation}
\label{eq-update}
{\bf{M}}_t = \left\{
\begin{array}{lcr}
{\varphi ({\bf{M}}_t^{'},{W_w}{{\bf{x}}_t})} &{{\bf{s}}_t = 1}\\
{\bf{M}}_t^{'} &{{\bf{s}}_t = 0}
\end{array} \right.,
\end{equation}
where ${W_w} \in {R^{D \times d}}$ is a matrix that transform convolutional feature ${\bf{x}}_t$ into memory item ${\bf{m}}_{N_t+1}$, and the write function $\varphi(\cdot)$ is implemented as list append operation. Obviously, the external memory module is updated only when ${{\bf{s}}_t = 1}$ and remain the same when ${{\bf{s}}_t = 0}$. With large-scale datasets, the model can learn to write relevant information into memory module and neglect noise from scratch, which is crucial to model complex motion dynamics in long-range temporal structure. Except memory history and convolutional feature, context embedding is another factor to determine the memory write decision, and we will introduce it in the next section.

\textbf{Context Embedding.} Due to LSTM's limited ability to address the non-stationary issue of long-term motion dynamics \cite{sun2017lattice}, we build a segmental recurrent cell on top of LSTM unit, which alleviates the non-stationary issue by split a long sequence into several short segments. At its core, there is a one-dimensional vector ${\bf{v}}_t \in \{0,1\}$ which indicates whether continue updating the LSTM's state at current timestep or reinitializes them to zeros. Because long-term motion dynamics usually shows segmental characteristics, in practice, we just replace this indicator vector with the memory discrete decision ${\bf{s}}_t$, i.e., ${\bf{v}}_t = {\bf{s}}_t$, which reduces the computation and makes sense to some extent. Specifically, indicator vector ${\bf{v}}_t$ decide whether to transfer the hidden state and memory cell content to the next timestep or to reinitialize them,
\begin{equation}
\label{eq-lstm}
{\hat{\bf{h}}}_t,{\hat{\bf{c}}}_t = LSTM({\bf{x}}_t;{\bf{h}}_{t - 1},{\bf{c}}_{t - 1}),
\end{equation}
\begin{equation}
\label{eq-update_h}
{\bf{h}}_t = {\bf{s}}_t  \times {\hat{\bf{h}}}_t,
\end{equation}
\begin{equation}
\label{eq-update_c}
{\bf{c}}_t = {\bf{s}}_t  \times {\hat{\bf{c}}}_t.
\end{equation}
The resulting state and memory are employed to compute gates values at the next time step. The LSTM breaks the connection with the previous hidden states ${\bf{h}}_{t-1}$, ${\bf{c}}_{t-1}$ and reinitializes them to zeros if ${\bf{s}}_t=0$, and the hidden state of timestep $t$ is passed to $t+1$ when ${\bf{s}}_t=1$. We use ${\hat{\bf{h}}}_t$ as the context embedding ${\bf{ce}}_t$ at time $t$.

\subsection{Discrete Memory Controller}
In previous works on memory networks, the memory controller is constructed by a feed-forward network or LSTM, which interacts with the external memory module using a number of read and write instructions. We employ the similar idea but the controller only outputs the write instruction that acts to place informative features into memory module, because the read process is performed at each timestep. Specifically, for each time step, the memory controller takes the convolutional feature ${\bf{x}}_t$, memory history ${\bf{mh}}_t$ and context embedding ${\bf{ce}}_t$ as inputs, and outputs the discrete memory write decision ${\bf{s}}_t \in \{0,1\}$. Formally, the write decision ${\bf{s}}_t$ is computed as a linear combination of these three inputs, followed by a function $\tau$ which is the composition of sigmoid function and a hard threshold function:
\begin{equation}
\label{eq-decision}
{\bf{q}}_t = {\bf{v}}_s^T \cdot ReLU({W_{sf}}{{\bf{x}}_t} + {W_{sc}}{{\bf{ce}}_t} + {W_{sm}}{{\bf{mh}}_t} + {{\bf{b}}_s}),
\end{equation}
\begin{equation}
\label{eq-decision}
{\bf{a}}_t = \sigma ({\bf{q}}_t),
\end{equation}
\begin{equation}
\label{eq-apply-step}
{\bf{s}}_t = \tau ({\bf{a}}_t),
\end{equation}
\begin{equation}
\label{eq-tau}
\tau (x) = \left\{
\begin{array}{lcr}
1, &if \; x > thr\\
0, &otherwise
\end{array} \right.,
\end{equation}
where ${\bf{v}}_s^T$ is learnable row vector and ${W_{sf}}$, ${W_{sc}}$, ${W_{sm}}$ and ${{\bf{b}}_s}$ are learned weights and biases. Firstly, sigmoid function $\sigma$ is applied on the outputs from linear combination to normalize ${\bf{a}}_t \in (0,1)$, which represents the importance of current frame for recognizing action in the video. Then, we adopt a simple threshold-based selection mechanism to decide whether write current convolutional feature ${\bf{x}}_t$ into external memory module, where $thr$ is a hyper-parameter and set to 0.5 in our experiments.

\textbf{Straight-Through Estimator.} The discrete decisions involved in transferring the binary variable ${\bf{s}}_t$ with the hard threshold function $\tau$ make it impossible to use standard gradient back-propagation to learn the parameters of the memory controller. To solve this issue, straight-through estimator proposed by \cite{bengio2013estimating} is employed to obtain to gradients of the memory controller. The idea is that a differentiable approximation function is used to substitute hard threshold function. In our case, we approximate the hard threshold function with the identity function, which has shown its effectiveness when considering a single layer of neurons. After replacing hard threshold function with identity function, the chain rule between ${\bf{s}}_t$ and ${\bf{q}}_t$ is
\begin{equation}
\begin{split}
\frac{{\partial {{\bf{s}}_t}}}{{\partial {{\bf{q}}_t}}} & = \frac{{\partial {{\bf{s}}_t}}}{{\partial {{\bf{a}}_t}}} \cdot \frac{{\partial {{\bf{a}}_t}}}{{\partial {{\bf{q}}_t}}} = 1 \cdot \frac{{\partial {{\bf{a}}_t}}}{{\partial {{\bf{q}}_t}}} \\
& = \sigma ({\bf{q}}_t)(1 - \sigma ({\bf{q}}_t)),
\end{split}
\label{eq-st}
\end{equation}
which enable the model can be trained in end-to-end manner. At inference time, the hard threshold function is used to output memory write signal. In this way, the discrepancy between training and test stage is eliminated in forward pass.

\begin{figure}
  \centering
  \includegraphics[width=.50\textwidth]{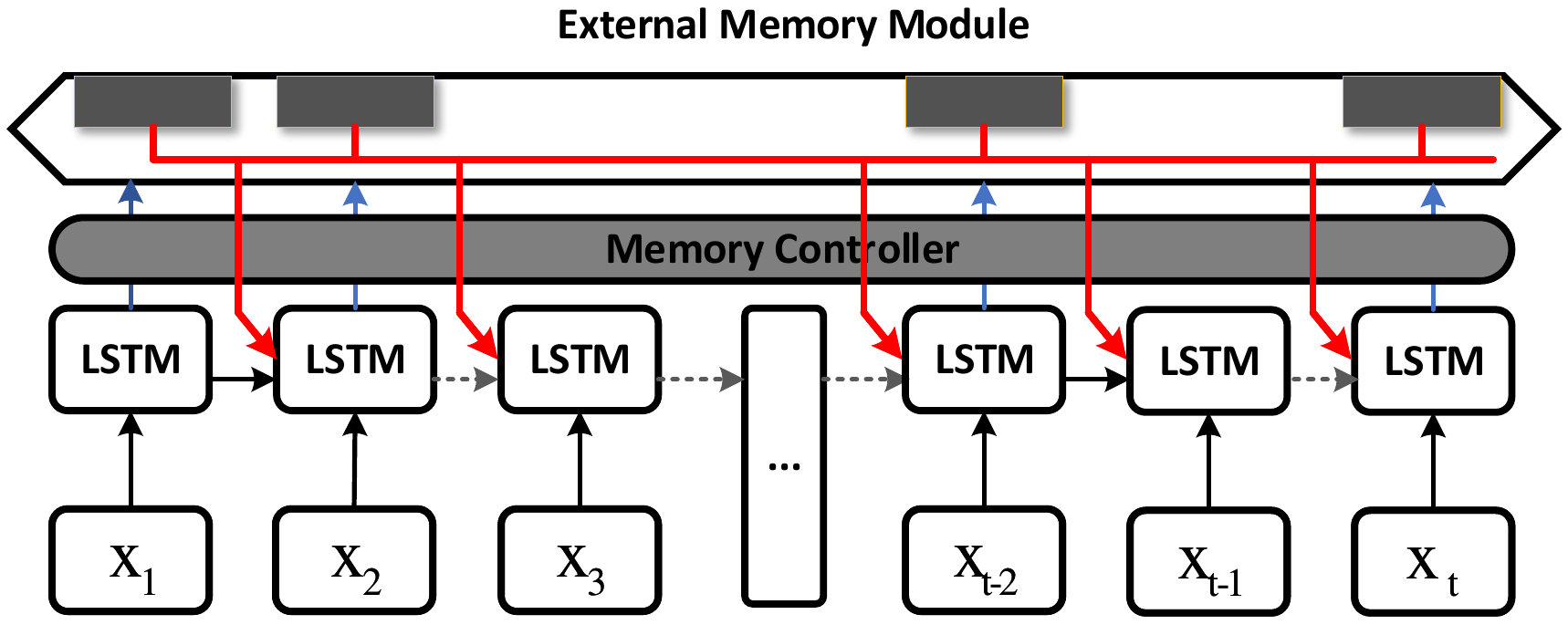}
  \caption{Illustration of the data flow in our proposed model. `LSTM' boxes represent the linear operations in Eq. \ref{eq-lstm}, the black solid line between `LSTM' boxes represents the hidden states update operation and gray dashed line means reinitialize hidden state to zero, as shown in Eq. \ref{eq-update_h}, Eq. \ref{eq-update_c}. Additionally, blue line cross memory controller box means that corresponding features are written into memory module and the red line represents the read processing of external memory module.}\label{Fig-data-flow}
\end{figure}

\subsection{Learning Architecture}
As illustrated in Figure \ref{Fig-overview}, the two-stream framework is adopted to improve the accuracy of prediction, where the optical flow is utilized as an additional modality to compensate the RGB inputs. The optical flow inputs are complementary to our proposed model, because optical flow only captures small motion between consecutive frames and our model aims to learn long-term motion patterns in long-range temporal structure. Note that video sequences, each composed of several frames sampled from a video, are fed into convolutional network to extract high-level feature representations. Therefore, our approach can be extendible for almost all CNN architectures, including BN-Inception, ResNets, and VGGnet. In practice, we initialize our convolutional network with models that are well pre-trained on ImageNet before we fine-tune them on the relatively small video datasets.

\textbf{Computation Graph Analysis.} From data flow perspective, our approach enables the model to reroute the forward path adaptively, which provides an effective and efficient way to exploit the temporal correlations in long-range motion dynamics. As shown in Figure \ref{Fig-data-flow}, suppose the convolutional feature $\bf{x}_t$ is written into external memory module, i.e., $\bf{s}_t=1$, and it will be connected with subsequent frames through memory read processing such as $\bf{x}_1$ and $\bf{x}_2$ in Figure \ref{Fig-data-flow}. Meanwhile, the inherent connection in LSTM unit is broken using Eq. \ref{eq-update_h} and Eq. \ref{eq-update_c} when $\bf{s}_t=0$, e.g. dashed line between $\bf{x}_2$ and $\bf{x}_3$ in the figure. Compared to traditional LSTM, all information from foregoing frames that are transformed into memory items are utilized to compute the gate value $\bf{s}_t$ rather than only last frame, which improves capacity for modeling long-term and complex motion dynamics. For example, the information of $\bf{x}_1$ and $\bf{x}_2$ is passed to frame $t-1$ and $t$ through ``red'' data path in Figure \ref{Fig-data-flow}. Moreover, all parameters in the memory controller are learned from video datasets and the data flow of forward pass varies in different video sequences by using sample-dependent variable $\bf{s}_t$.

\section{Experiments}
To evaluate the proposed network architecture, we conduct action recognition experiments on two benchmark datasets, with in-depth comparisons with baseline and other architectures to verify our design principles. We also provide visualization of the learned memory controller as a critic part of the model's interpretability.

\subsection{Experimental Settings}
\textbf{Dataset.} Experiments are mainly conducted on two action recognition benchmark datasets: UCF101 \cite{soomro2012ucf101} and HMDB51 \cite{kuehne2011hmdb}. The UCF101 dataset is composed of 13,320 videos categorized into 101 actions, ranging from daily life activities to unusual sports. The videos are collected from Internet with various camera motions and illuminations. HMDB51 dataset contains 6766 videos divided into 51 action classes. Complex backgrounds and similar scenes in different categories make this dataset more challenging than others. For both of them, we follow the provided evaluation protocol and adopt standard training/test splits and report the mean classification accuracy over these splits.

\textbf{Implementation Details.} We choose VGG16 and ResNet101 as our CNN feature extractor for the RGB and optical flow images. We use stochastic gradient descent algorithm to train our model from scratch. The mini-batch size is set to 64 and the momentum is set to 0.9. We use small learning rate in our experiments. For spatial-stream networks, the learning rate is initialized as 0.001 and decrease by $\frac{1}{{10}}$ every 6,000 iterations. The training procedure stops after 18,000 iterations. For the temporal stream, we initialize the learning rate as 0.005, which reduces to its $\frac{1}{{10}}$ after 48,000 and 72,000 iterations. The maximum iteration is set as 80,000. We use gradient clipping of 20 to avoid exploding gradient at the early stage. For data augmentation, we use the techniques of location jittering, horizontal flipping, corner cropping, and scale jittering, as described in \cite{wang2016temporal}. All the experiments are run on the PyTorch toolbox \cite{paszke2017automatic}.

\textbf{Baseline Two-stream ConvNets.} There are two types of two-stream baseline. One is the convNet-baseline reported in \cite{li2018videolstm} and another one is TSN-baseline proposed in \cite{wang2016temporal}. We distinguish these two baseline-based method by w/o ``+TSN'', i.e., Ours(VGG)/Ours(VGG+TSN) and Ours(ResNet+TSN).

\begin{table}[htbp]
\centering
\caption{Performance evaluation on videos of split 1 of UCF-101 with complex movements.}
\begin{tabular}{c|c|c}
\hline
Action Category  &ConvLSTM &Ours(VGG)\\
\hline
Human-Object Interaction &76.0 &83.3\\
Human-Human Interaction &89.1 &91.5\\
Body Motion Only &88.1 &89.3\\
\hline
\hline
Pizza Tossing &21.1 &54.6\\
Mixing Batter &55.6 &75.6\\
Playing Dhol &81.6 &95.9\\
Salsa Spins &86.0 &93.1\\
Ice Dancing &97.8 &98.0\\
\hline
\end{tabular}\label{Table-complex-movements}
\end{table}

\begin{figure*}
\begin{center}
   \includegraphics[width=1.0\linewidth]{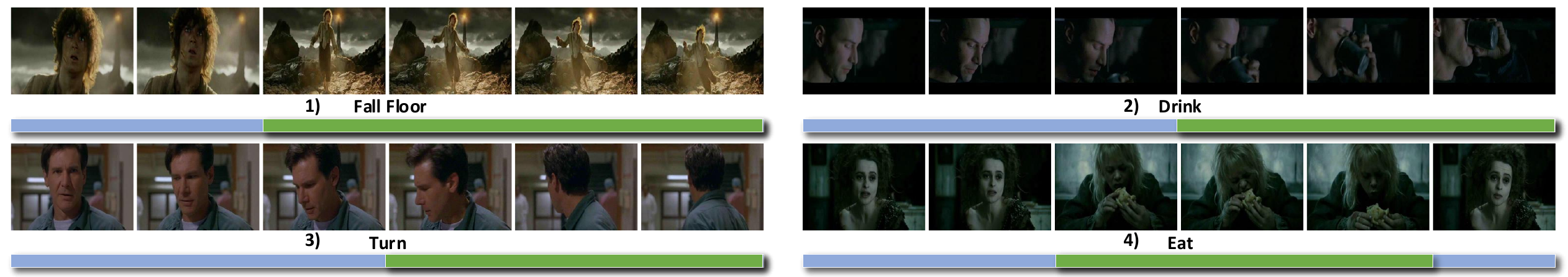}
\end{center}
     \caption{Memory controller decision results of the proposed method on HMDB51 dataset. Note that the memory controller can correctly select the most evident frames and ignore irrelevant frames. Specifically, the blue bar stands for the memory controller decision ${\bf{s}_t=0}$, and the green bar stands for the memory controller decision ${\bf{s}_t=1}$, i.e., corresponding convolutional features are written into external memory module.}
\label{Fig-mc-visu}
\end{figure*}

\subsection{Effect on videos with complex movements}
In this subsection, We focus on the effectiveness of our model processing videos with complex motion dynamics. The 101 action categories of UCF101 dataset have been divided into five coarse types: human-object interaction, body-motion only, human-human interaction, playing musical instruments, and sports. We summarize the classification accuracy on these coarse categories and report it in Table \ref{Table-complex-movements}. The upper part of the table shows the performance on the coarse category. The ConvLSTM that aims to model spatio-temporal pattern in complex motion dynamics are utilized for comparison, and our model outperforms on all coarse categories, particularly by 7.3\% on human-object interaction, by 2.4\% on human-human interaction and 1.2\% on body motion only. The impressive improvements on human interacting actions verify our model's superiority in handling complex motion dynamics over ConvLSTM. Meanwhile, our method shows similar results with ConvLSTM on body motion only actions since the movements are simple in this category. Furthermore, results from several typical classes are shown in lower part of the table. Our model gains 33.5\% and 20.0\% improvements on pizza tossing and mixing batter category respectively, where the movements between the person and object are complex and fast. In the ice dancing category, we see only 0.2\% improvement since the appearance information is very discriminative and recognition accuracy is comparable using one-single frame.

\begin{table}[htbp]
\centering
\caption{Detailed results on videos of split 1 of UCF-101 using VGG16.}
\begin{tabular}{c|c|c}
\hline
Method  &Spatial Networks &Temporal Networks\\
\hline
ConvNet &77.4 &75.2\\
LSTM &77.5 &78.3\\
ALSTM &77.0 &79.5\\
ConvLSTM &77.6 &79.1\\
VideoLSTM &79.6 &82.1\\
\hline
\hline
Ours(VGG) &80.0 &82.7\\
\hline
\end{tabular}\label{Table-ucf101-split1}
\end{table}

\subsection{Memory Controller Visualization}
We present several video sequences to visually verify the effect of the discrete memory controller, which is the key component that rules the memory content for action recognition. Figure \ref{Fig-mc-visu} shows the memory write decisions and corresponding RGB frame on split 1 of HMDB51 datasets. We can clearly see that the memory controller learns to write the most evident information into external memory module and ignore irrelevant frames.

In Figure \ref{Fig-mc-visu}, we show four video sequences sampled from fall floor, drink, eat, and turn categories. These actions have strong segmental characteristics, for example, drink can be divided into three stages: grab the glass, drink and put down the glass. And the learned memory controller can select the most discriminative stage to recognize actions. For instance, in sequence 1), there is a close-up of the person's head in the beginning which is irrelevant to the fall floor activity, so the memory controller decides to ignore these close-up frames and store the relevant frames once they appear. What is more, in sequence 2) and 3), drink and turn activities are composed of several stages and the activities are usually confusing before real actions occur. For example, the person may put down the glass instead of drink it after observed three frames in sequence 2). Therefore, the learned memory controller tends to write the most evident frames into the external memory module even though losing some relevant information. At the same time, the learned memory controller is robust to background and object variation. The changes of person and background make the sequence 4) more challenging, and our controller can autonomically find the leading person and ignore others.

\begin{table*}[htbp]
\centering
\caption{State-of-the-art comparison with LSTM-like architectures.}
\begin{tabular}{c|cccccccc}
\hline
\multirow{2}{*}{Method} &
\multicolumn{2}{c} {Pre-training} &
\multicolumn{2}{c} {Fusion} &
\multicolumn{2}{c} {Networks} &
\multirow{2}{*}{UCF101} &
\multirow{2}{*}{HMDB51} \\
&ImageNet &1M Sports &Average &Product &VGG &ResNet \\
\hline
ConvNet &\checkmark &- &\checkmark &- &\checkmark &- &77.4 &75.2\\
ALSTM &\checkmark &- &\checkmark &- &\checkmark &- &77.0 &41.3\\
VideoLSTM &\checkmark &- &- &\checkmark  &\checkmark &- &89.2 &56.4\\
L$^2$STM &\checkmark &- &\checkmark &- &\checkmark &- &93.6 &66.2\\
TSN(ResNet) &\checkmark &- &\checkmark &- &- &\checkmark &93.9 &69.7\\
TS-LSTM &\checkmark &- &\checkmark &- &- &\checkmark &94.1 &69.0\\
\hline
\hline
Ours(VGG+TSN) &\checkmark &- &\checkmark &- &\checkmark &- &92.1 &67.3\\
Ours(ResNet+TSN) &\checkmark &- &\checkmark &- &- &\checkmark &94.8 &71.8\\
\hline
\end{tabular}\label{Table-all-lstm}
\end{table*}

\subsection{Comparison with other temporal models}
Recurrent neural network, especially LSTM, have the ability to model temporal motion dynamics, and shown moderate improvement from previous work. Moreover, there are several LSTM variants aiming at learning complex motion dynamics from video, such as ConvLSTM, soft-Attention LSTM (ALSTM) and VideoLSTM \cite{li2018videolstm}. Due to the VGG16 convolutional network are used as feature extractor in these methods, we choose VGG16 network as base architecture and list detailed results in Table \ref{Table-ucf101-split1}. The vanilla LSTM only obtains no improvement on spatial network and 3.1\% improvement on temporal network. By enhancing the network with the external memory module, our model produces the highest accuracy in both stream network and improves the performance by 2.6 and 7.5 points respectively. Furthermore, our model also outperforms other LSTM variants that are designed to action recognition.

In Table \ref{Table-all-lstm}, we list all state-of-the-art methods which learn motion patterns with LSTM for action recognition tasks. In order to present the completely and clearly, we elaborate on some factors, such as pre-training type, fusion strategy, and network architectures. To be specific, we compare the proposed method with ALSTM \cite{sharma2015action}, VideoLSTM \cite{li2018videolstm}, L$^2$STM \cite{sun2017lattice} and TS-LSTM \cite{ma2017ts}. The recent state-of-the-art method L$^2$STM and TS-LSTM also aim at modeling long-term motion dynamics in the videos, and our method outperforms L$^2$STM in terms of average accuracy 67.3\% vs 66.2\% on HMDB51 dataset and obtains comparable performance on UCF101 dataset. The reason behind this is the videos from UCF101 are trimmed carefully, which makes our method less useful, and a cross-modal training strategy is utilized to boost final performance. Compared with UCF101 dataset, videos from HMDB51 dataset usually contains some action-irrelevant shot, as shown in Figure \ref{Fig-mc-visu}, which makes HMDB51 dataset more challenging. And our proposed model tackles these video better by storing useful information and ignore noisy feature representations, so our approach makes a consistent improvement on HMDB51 dataset when compared with other methods.

\subsection{Comparison with the state-of-the-art}
In addition to the various temporal models comparison in Table \ref{Table-all-lstm}, we compare our method with other state-of-the-art algorithms and the results are reported in Table \ref{Table-final}. We evaluate our method following the testing scheme described in the standard two-stream method \cite{simonyan2014two}, where final classification results are obtained by average of the spatial and temporal stream results. Specifically, we compare our method with traditional approaches such as improved trajectories (iDTs) \cite{wang2011action} and deep learning algorithms such as two-stream networks \cite{simonyan2014two}, factorized spatio-temporal convolutional networks (F$_{ST}$CN) \cite{sun2015human}, 3D convolutional networks (C3D) \cite{tran2015learning}, spatio-temporal fusion CNNs \cite{feichtenhofer2016convolutional}, attention cluster (Att-C) \cite{long2018attention} and TSN \cite{wang2016temporal}, which model long-term temporal structure by segmental architecture with sparse sampling.

As shown in Table \ref{Table-final}, our best implementation based on ResNet improves the average accuracy by 0.7\% on UCF101 and 2.5\% on HMDB51, which is reported in TS-LSTM using the same ResNet architecture. For testifying that our method is generally effective, we additionally evaluate our model with VGG-16 network as the previous two-stream CNNs architectures \cite{simonyan2014two}. Both based on VGG-16, our method (92.1\%) is still comparable to the spatio-temporal fusion CNNs (92.5\%) \cite{feichtenhofer2016convolutional} on UCF101 datasets and outperforms it by 1.9\% on HMDB51 dataset. Meanwhile, our method achieves the best results 94.8\% and 71.8\% on UCF101 and HMDB51 dataset respectively. This result demonstrates that our approaches can be widely applied to many fancy CNN models.

\begin{table}[htbp]
\centering
\caption{Performance comparison with the state-of-the-art.}
\begin{tabular}{|c|c|c|c|}
\hline
Model &Method &UCF101 &HMDB51\\
\hline
\multirow{4}{*}{Traditional}
&iDT + FV &85.9 &57.2\\
&iDT + HSV &87.9 &61.1\\
&VideoDarwin &- &63.7\\
&MPR &- &65.5\\
\hline
\hline
\multirow{3}{*}{Deep}
&Two Stream &88.0 &59.4\\
&F$_{ST}$CN &88.1 &59.1\\
&VideoLSTM &89.2 &56.4\\
\hline
\hline
\multirow{5}{*}{Very Deep}
&C3D &85.2 &-\\
&Fusion &92.5 &65.4\\
&L$^2$STM &93.6 &66.2\\
&TS-LSTM &94.1 &69.0\\
&TSN(BN-Inception) &94.0 &68.5\\
&TSN(ResNet) &93.9 &69.7\\
&Att-C &94.6 &69.2\\
\hline
\hline
\multirow{2}{*}{Ours}
&Ours(VGG16+TSN) &92.1 &67.3\\
&Ours(ResNet+TSN) &\bf{94.8} &\bf{71.8}\\
\hline
\end{tabular}\label{Table-final}
\end{table}

\section{Conclusion}
In this paper, we present a memory-augmented temporal dynamic learning network to address the challenging problem of modeling long-term and complex motion dynamics. Particularly, we attach a simple and growing external memory module to CNN to store the most evident and relevant information for recognizing action in the video. Furthermore, through memory items in the external memory module, the model can route the forward path adaptively using the discrete memory controller. As the core of the network, the memory controller employs hard threshold function to output discrete memory write decision and approximates its gradients using straight-through estimator, which enables the network can be trained end-to-end from scratch. As shown in visualization results, the learned memory controller can write the most evident information into external memory module and ignore irrelevant frames automatically. Moreover, to verify the generalization ability of the proposed model, we evaluate our model with different CNN architecture. Experiments conducted on both UCF101 and HMDB51 datasets validate our proposal and analysis.

\section{Acknowledgments}
This work was supported by the National Key R\&D Program of China under Grant 2017YFB1002202, National Natural Science Foundation of China under Grant 61773316, Natural Science Foundation of Shaanxi Province under Grant 2018KJXX-024, Fundamental Research Funds for the Central Universities under Grant 3102017AX010, and the Open Research Fund of Key Laboratory of Spectral Imaging Technology, Chinese Academy of Sciences.

\bibliographystyle{aaai}
\bibliography{AAAI-Paper}

\begin{thebibliography}{}

\bibitem[\protect\citeauthoryear{Bengio, L{\'e}onard, and
  Courville}{2013}]{bengio2013estimating}
Bengio, Y.; L{\'e}onard, N.; and Courville, A.
\newblock 2013.
\newblock Estimating or propagating gradients through stochastic neurons for
  conditional computation.
\newblock {\em arXiv preprint arXiv:1308.3432}.

\bibitem[\protect\citeauthoryear{Boiman and Irani}{2007}]{boiman2007detecting}
Boiman, O., and Irani, M.
\newblock 2007.
\newblock Detecting irregularities in images and in video.
\newblock {\em International journal of computer vision} 74(1):17--31.

\bibitem[\protect\citeauthoryear{Carreira and
  Zisserman}{2017}]{carreira2017quo}
Carreira, J., and Zisserman, A.
\newblock 2017.
\newblock Quo vadis, action recognition? a new model and the kinetics dataset.
\newblock In {\em IEEE Conference on Computer Vision and Pattern Recognition},
  4724--4733.

\bibitem[\protect\citeauthoryear{Feichtenhofer, Pinz, and
  Wildes}{}]{feichtenhofer2017multiplier}
Feichtenhofer, C.; Pinz, A.; and Wildes, R.~P.
\newblock Spatiotemporal multiplier networks for video action recognition.

\bibitem[\protect\citeauthoryear{Feichtenhofer, Pinz, and
  Zisserman}{2016}]{feichtenhofer2016convolutional}
Feichtenhofer, C.; Pinz, A.; and Zisserman, A.
\newblock 2016.
\newblock Convolutional two-stream network fusion for video action recognition.
\newblock In {\em IEEE Conference on Computer Vision and Pattern Recognition},
  1933--1941.

\bibitem[\protect\citeauthoryear{Gao \bgroup et al\mbox.\egroup
  }{2018}]{gao2018motion}
Gao, J.; Ge, R.; Chen, K.; and Nevatia, R.
\newblock 2018.
\newblock Motion-appearance co-memory networks for video question answering.
\newblock {\em arXiv preprint arXiv:1803.10906}.

\bibitem[\protect\citeauthoryear{Graves, Wayne, and
  Danihelka}{2014}]{graves2014neural}
Graves, A.; Wayne, G.; and Danihelka, I.
\newblock 2014.
\newblock Neural turing machines.
\newblock {\em arXiv preprint arXiv:1410.5401}.

\bibitem[\protect\citeauthoryear{Joulin and
  Mikolov}{2015}]{joulin2015inferring}
Joulin, A., and Mikolov, T.
\newblock 2015.
\newblock Inferring algorithmic patterns with stack-augmented recurrent nets.
\newblock In {\em Advances in Neural Information Processing Systems},
  190--198.

\bibitem[\protect\citeauthoryear{Karpathy \bgroup et al\mbox.\egroup
  }{2014}]{karpathy2014large}
Karpathy, A.; Toderici, G.; Shetty, S.; Leung, T.; Sukthankar, R.; and Fei-Fei,
  L.
\newblock 2014.
\newblock Large-scale video classification with convolutional neural networks.
\newblock In {\em IEEE Conference on Computer Vision and Pattern Recognition},
  1725--1732.

\bibitem[\protect\citeauthoryear{Krizhevsky, Sutskever, and
  Hinton}{2012}]{krizhevsky2012imagenet}
Krizhevsky, A.; Sutskever, I.; and Hinton, G.~E.
\newblock 2012.
\newblock Imagenet classification with deep convolutional neural networks.
\newblock In {\em Advances in Neural Information Processing Systems},
  1097--1105.

\bibitem[\protect\citeauthoryear{Kuehne \bgroup et al\mbox.\egroup
  }{2011}]{kuehne2011hmdb}
Kuehne, H.; Jhuang, H.; Garrote, E.; Poggio, T.; and Serre, T.
\newblock 2011.
\newblock Hmdb: a large video database for human motion recognition.
\newblock In {\em IEEE International Conference on Computer Vision},
  2556--2563.

\bibitem[\protect\citeauthoryear{Li \bgroup et al\mbox.\egroup
  }{2018}]{li2018videolstm}
Li, Z.; Gavrilyuk, K.; Gavves, E.; Jain, M.; and Snoek, C.~G.
\newblock 2018.
\newblock Videolstm convolves, attends and flows for action recognition.
\newblock {\em Computer Vision and Image Understanding} 166:41--50.

\bibitem[\protect\citeauthoryear{Lin \bgroup et al\mbox.\egroup
  }{2008}]{lin2008human}
Lin, W.; Sun, M.-T.; Poovandran, R.; and Zhang, Z.
\newblock 2008.
\newblock Human activity recognition for video surveillance.
\newblock In {\em International Symposium on Circuits and Systems},
  2737--2740.

\bibitem[\protect\citeauthoryear{Long \bgroup et al\mbox.\egroup
  }{2018a}]{long2018multimodal}
Long, X.; Gan, C.; de~Melo, G.; Liu, X.; Li, Y.; Li, F.; and Wen, S.
\newblock 2018a.
\newblock Multimodal keyless attention fusion for video classification.
\newblock In {\em AAAI Conference on Artificial Intelligence}.

\bibitem[\protect\citeauthoryear{Long \bgroup et al\mbox.\egroup
  }{2018b}]{long2018attention}
Long, X.; Gan, C.; de~Melo, G.; Wu, J.; Liu, X.; and Wen, S.
\newblock 2018b.
\newblock Attention clusters: Purely attention based local feature integration
  for video classification.
\newblock In {\em IEEE Conference on Computer Vision and Pattern Recognition},
  7834--7843.

\bibitem[\protect\citeauthoryear{Long, Shelhamer, and
  Darrell}{2015}]{long2015fully}
Long, J.; Shelhamer, E.; and Darrell, T.
\newblock 2015.
\newblock Fully convolutional networks for semantic segmentation.
\newblock In {\em IEEE Conference on Computer Vision and Pattern Recognition},
  3431--3440.

\bibitem[\protect\citeauthoryear{Ma \bgroup et al\mbox.\egroup
  }{2017}]{ma2017ts}
Ma, C.-Y.; Chen, M.-H.; Kira, Z.; and AlRegib, G.
\newblock 2017.
\newblock Ts-lstm and temporal-inception: Exploiting spatiotemporal dynamics
  for activity recognition.
\newblock {\em arXiv preprint arXiv:1703.10667}.

\bibitem[\protect\citeauthoryear{Paszke \bgroup et al\mbox.\egroup
  }{2017}]{paszke2017automatic}
Paszke, A.; Gross, S.; Chintala, S.; Chanan, G.; Yang, E.; DeVito, Z.; Lin, Z.;
  Desmaison, A.; Antiga, L.; and Lerer, A.
\newblock 2017.
\newblock Automatic differentiation in pytorch.
\newblock In {\em Advances in Neural Information Processing Systems Workshop}.

\bibitem[\protect\citeauthoryear{Sharma, Kiros, and
  Salakhutdinov}{2015}]{sharma2015action}
Sharma, S.; Kiros, R.; and Salakhutdinov, R.
\newblock 2015.
\newblock Action recognition using visual attention.
\newblock {\em arXiv preprint arXiv:1511.04119}.

\bibitem[\protect\citeauthoryear{Simonyan and
  Zisserman}{2014}]{simonyan2014two}
Simonyan, K., and Zisserman, A.
\newblock 2014.
\newblock Two-stream convolutional networks for action recognition in videos.
\newblock In {\em Advances in Neural Information Processing Systems},
  568--576.

\bibitem[\protect\citeauthoryear{Soomro, Zamir, and
  Shah}{2012}]{soomro2012ucf101}
Soomro, K.; Zamir, A.~R.; and Shah, M.
\newblock 2012.
\newblock Ucf101: A dataset of 101 human actions classes from videos in the
  wild.
\newblock {\em arXiv preprint arXiv:1212.0402}.

\bibitem[\protect\citeauthoryear{Sukhbaatar \bgroup et al\mbox.\egroup
  }{2015}]{sukhbaatar2015end}
Sukhbaatar, S.; Weston, J.; Fergus, R.; et~al.
\newblock 2015.
\newblock End-to-end memory networks.
\newblock In {\em Advances in Neural Information Processing Systems},
  2440--2448.

\bibitem[\protect\citeauthoryear{Sun \bgroup et al\mbox.\egroup
  }{2015}]{sun2015human}
Sun, L.; Jia, K.; Yeung, D.-Y.; and Shi, B.~E.
\newblock 2015.
\newblock Human action recognition using factorized spatio-temporal
  convolutional networks.
\newblock In {\em IEEE International Conference on Computer Vision},
  4597--4605.

\bibitem[\protect\citeauthoryear{Sun \bgroup et al\mbox.\egroup
  }{2017}]{sun2017lattice}
Sun, L.; Jia, K.; Chen, K.; Yeung, D.-Y.; Shi, B.~E.; and Savarese, S.
\newblock 2017.
\newblock Lattice long short-term memory for human action recognition.
\newblock In {\em IEEE International Conference on Computer Vision},
  2166--2175.

\bibitem[\protect\citeauthoryear{Tran \bgroup et al\mbox.\egroup
  }{2015}]{tran2015learning}
Tran, D.; Bourdev, L.; Fergus, R.; Torresani, L.; and Paluri, M.
\newblock 2015.
\newblock Learning spatiotemporal features with 3d convolutional networks.
\newblock In {\em IEEE International Conference on Computer Vision},
  4489--4497.

\bibitem[\protect\citeauthoryear{Vu \bgroup et al\mbox.\egroup
  }{2018}]{vu2018memory}
Vu, T.-H.; Choi, W.; Schulter, S.; and Chandraker, M.
\newblock 2018.
\newblock Memory warps for learning long-term online video representations.
\newblock {\em arXiv preprint arXiv:1803.10861}.

\bibitem[\protect\citeauthoryear{Wang \bgroup et al\mbox.\egroup
  }{2011}]{wang2011action}
Wang, H.; Kl{\"a}ser, A.; Schmid, C.; and Liu, C.-L.
\newblock 2011.
\newblock Action recognition by dense trajectories.
\newblock In {\em IEEE Conference on Computer Vision and Pattern Recognition},
  3169--3176.

\bibitem[\protect\citeauthoryear{Wang \bgroup et al\mbox.\egroup
  }{2016}]{wang2016temporal}
Wang, L.; Xiong, Y.; Wang, Z.; Qiao, Y.; Lin, D.; Tang, X.; and Van~Gool, L.
\newblock 2016.
\newblock Temporal segment networks: Towards good practices for deep action
  recognition.
\newblock In {\em European Conference on Computer Vision},  20--36.

\bibitem[\protect\citeauthoryear{Wang \bgroup et al\mbox.\egroup
  }{2018}]{8418840}
Wang, Q.; Yuan, Z.; Du, Q.; and Li, X.
\newblock 2018.
\newblock Getnet: A general end-to-end 2-d cnn framework for hyperspectral
  image change detection.
\newblock {\em IEEE Transactions on Geoscience and Remote Sensing}.

\bibitem[\protect\citeauthoryear{Xingjian \bgroup et al\mbox.\egroup
  }{2015}]{xingjian2015convolutional}
Xingjian, S.; Chen, Z.; Wang, H.; Yeung, D.-Y.; Wong, W.-K.; and Woo, W.-c.
\newblock 2015.
\newblock Convolutional lstm network: A machine learning approach for
  precipitation nowcasting.
\newblock In {\em Advances in Neural Information Processing Systems},
  802--810.

\bibitem[\protect\citeauthoryear{Yang and Chan}{2018}]{yang2018learning}
Yang, T., and Chan, A.~B.
\newblock 2018.
\newblock Learning dynamic memory networks for object tracking.
\newblock {\em arXiv preprint arXiv:1803.07268}.

\bibitem[\protect\citeauthoryear{Yue-Hei~Ng \bgroup et al\mbox.\egroup
  }{2015}]{yue2015beyond}
Yue-Hei~Ng, J.; Hausknecht, M.; Vijayanarasimhan, S.; Vinyals, O.; Monga, R.;
  and Toderici, G.
\newblock 2015.
\newblock Beyond short snippets: Deep networks for video classification.
\newblock In {\em IEEE Conference on Computer Vision and Pattern Recognition},
  4694--4702.

\bibitem[\protect\citeauthoryear{Zha \bgroup et al\mbox.\egroup
  }{2015}]{zha2015exploiting}
Zha, S.; Luisier, F.; Andrews, W.; Srivastava, N.; and Salakhutdinov, R.
\newblock 2015.
\newblock Exploiting image-trained cnn architectures for unconstrained video
  classification.
\newblock {\em arXiv preprint arXiv:1503.04144}.

\end{thebibliography}

\end{document}